\documentclass{article}
\usepackage{spconf,amsmath,epsfig}
\usepackage{color}
\let\OLDthebibliography\thebibliography
\renewcommand\thebibliography[1]{
  \OLDthebibliography{#1}
  \setlength{\parskip}{0pt}
  \setlength{\itemsep}{0pt plus 0.3ex}
}

\usepackage{amssymb,amsmath}
\begin{document}\sloppy

\def\x{{\mathbf x}}
\def\L{{\cal L}}

\title{Weakly supervised video anomaly detection based on cross-batch clustering guidance}
%
\name{Congqi Cao, Xin Zhang, Shizhou Zhang, Peng Wang, and Yanning Zhang}
\address{ASGO National Engineering Laboratory, School of Computer Science\\ 
Northwestern Polytechnical University\\
\{congqi.cao, szzhang, peng.wang, ynzhang\}@nwpu.edu.cn; zhangxin\_@mail.nwpu.edu.cn}

\maketitle

\begin{abstract}
Weakly supervised video anomaly detection (WSVAD) is a challenging task since only video-level labels are available for training. In previous studies, the discriminative power of the learned features is not strong enough, and the data imbalance resulting from the mini-batch training strategy is ignored. To address these two issues, we propose a novel WSVAD method based on cross-batch clustering guidance. To enhance the discriminative power of features, we propose a batch clustering based loss to encourage a clustering branch to generate distinct normal and abnormal clusters based on a batch of data. Meanwhile, we design a cross-batch learning strategy by introducing clustering results from previous mini-batches to reduce the impact of data imbalance. In addition, we propose to generate more accurate segment-level anomaly scores based on batch clustering guidance further improving the performance of WSVAD. Extensive experiments on two public datasets demonstrate the effectiveness of our approach.
\end{abstract}
\section{Introduction}
\label{sec:intro}

\begin{figure*}[htp]
	\centering
	\includegraphics[width=1.0\linewidth]{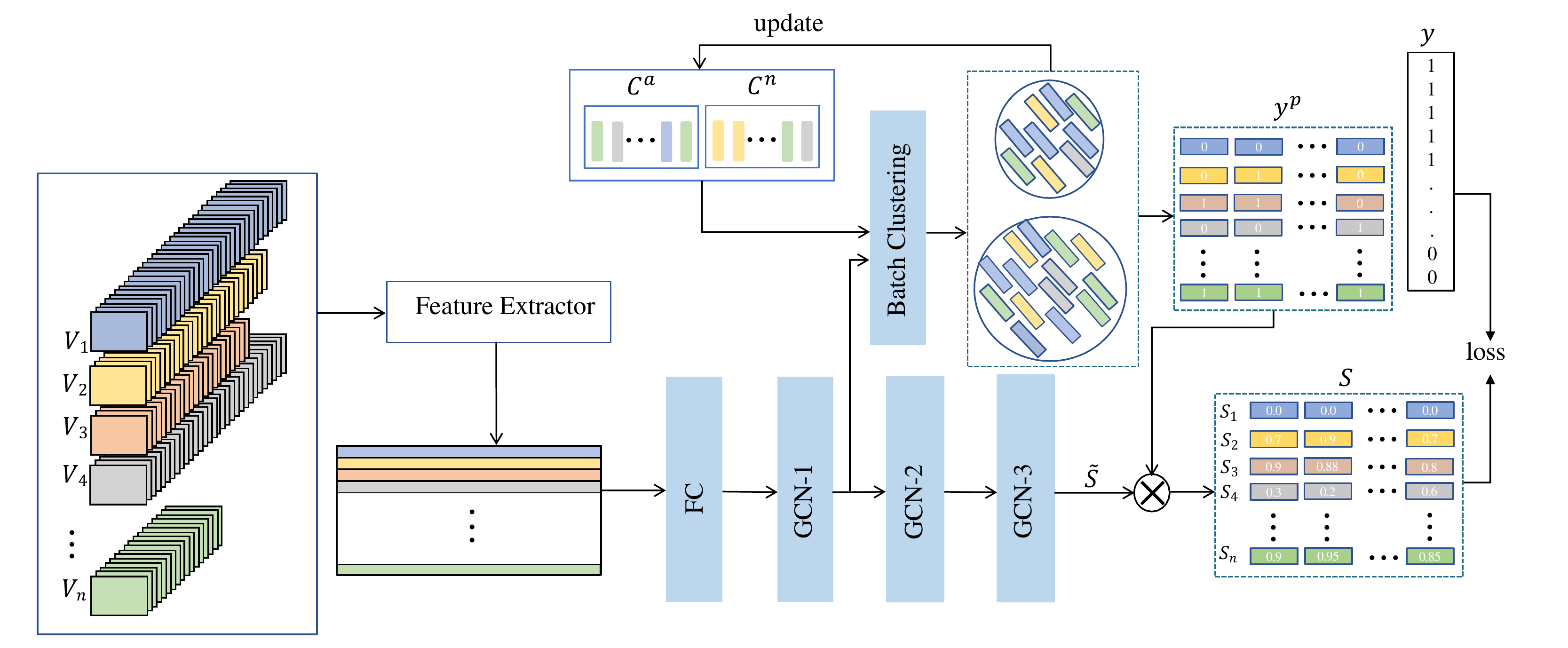}
	\caption{Overview of our proposed method. The feature extractor extracts features from video segments. The extracted features are fed into a fully connected layer and three graph convolution layers to generate segment-level anomaly scores. Simultaneously, the batch clustering branch uses intermediate representations of a batch of videos learned from the GCN-1 layer to create clusters. And we design a cross-batch learning strategy to store the clustering results of previous batches and introduce them into the batch clustering of the current batch. Finally, the pseudo labels generated by the batch clustering branch guide the generation of anomaly scores.}
	\label{fig:model}
\end{figure*}

An efficient and accurate video anomaly detection algorithm can help maintain social security and stability. Therefore, video anomaly detection has high practical value and broad application prospects. With the development of weakly supervised learning, the weakly supervised video anomaly detection (WSVAD) method is an effective method for detecting anomalies, which uses weakly labeled training data containing both normal and anomalous videos to train the model. Recently, WSVAD has been formulated as a multiple instance learning (MIL) task. Sultani et al. \cite{sultani2018real} constructed a large-scale anomaly dataset and proposed a deep MIL ranking based method for WSVAD. Wan et al. \cite{wan2020weakly} replaced the max anomaly score selection policy with a k-max value selection policy. Li et al. \cite{li2022self} selected the sequence with the highest sum of anomaly scores instead of selecting the instance with the highest anomaly score. Gong et al. \cite{gong2022multi} introduced temporal continuity of multiple neighboring instances at different time scales.

However, most of the existing work \cite{sultani2018real}\cite{wan2020weakly}\cite{li2022self} have only used MIL-based classification loss. Although MIL-based classification loss ensures the inter-class separability of the learned features to some extent, it is not sufficient for accurate anomaly detection. Therefore, we propose a batch clustering based loss to further increase the discriminative power of the features, and our framework is shown in Figure \ref{fig:model}. The abnormal videos in a batch are clustered into two clusters, and then the loss encourages the network to maximize the distance between these two clusters. Meanwhile, for normal videos in a batch, the loss encourages the network to minimize the distance between these two clusters. Compared to clustering each video individually \cite{zaheer2020self}\cite{zaheer2020claws}, batch clustering improves the robustness of the model and reduces the influence of the model affected by the noise.

Considering the data in WSVAD task is highly unbalanced, it may negatively affect model training when only using a small portion of data. Previous mini-batches can provide valuable knowledge enabling the model to better understand the underlying distribution of the data \cite{wang2020cross}\cite{wu2019long}. Therefore, we use a cross-batch learning strategy to provide guidance for the current batch clustering by introducing clustering results from previous batches. The introduced cross-batch learning strategy can make the clustering results more accurate, model the temporal-spatial distribution of the data better, and improve the adaptability of the model to unbalanced samples.

In addition, the knowledge that the model can learn in WSVAD directly from supervised learning is limited, since only video-level labels are available. The clustering branch can obtain potential information reflecting the similarity of video segments. Therefore, we propose anomaly score generation based on batch clustering guidance to generate more discriminative anomaly scores and further improve the model performance. We first generate pseudo labels for video segments based on the batch clustering results, and then uses the pseudo labels to guide the backbone network to rectify the estimated anomaly scores for video segments. In summary, our main contributions are as follows:

1) We propose a loss based on batch clustering to complement and enhance the separability of the features guided by MIL-based classification loss.

2) We propose a cross-batch learning strategy to generate more accurate clustering results.

3) We propose to use the knowledge learned from batch clustering to guide the predition of more discriminative anomaly scores.

\section{Proposed Method}
\label{method}
\subsection{Backbone network}
Our approach employs a backbone netwwork based on graph convolutional neural network (GCN) to model video sequences. The backbone network consists of a feature extraction module and a graph convolution module \cite{cao2022adaptive}. The Inflated 3D (I3D) \cite{carreira2017quo} pretrained on the Kinetics dataset is used as the feature extraction network to extract the appearance and motion information of the video segments. Before each video $V_i$ is fed into the feature extraction module, the video is divided into non-overlapping segments containing 16 consecutive frames, and we denote the number of segments by $T_i$. The graph convolution module consists three graph convolution layers, where the first two layers are followed by a ReLU activation function and a dropout layer, and the last layer is followed by a Sigmoid activation function. For each video $V_i$, the input layer receives the temporal-spatial features extracted from the feature extraction module and the adjacency matrix of a global graph constructed based on feature similarity and temporal proximity of the video segments. The output layer produces the anomaly score vector of the video $\tilde{S}_{i}=\left\{\tilde{s}_{i, j}\right\}_{j=1}^{T_{i}}$. The network is trained using video-level labels. $y_{i} \in\{0,1\}$ is the video-level label of video $V_i$, where $y_i=0$ indicates that video $V_i$ is a normal video and $y_i=1$ indicates that $V_i$ is an abnormal video.

\subsection{Batch clustering based on K-means}
Although MIL-based classification loss ensures the inter-class separability of learned features to a certain extent, it cannot ensure a more discriminative power of features since there is no explicit supervision. Several studies on unsupervised anomaly detection \cite{perera2019learning} have enlightened us, in which normal samples are compulsorily clustered in a compact space such that they can be kept away from the anomaly space. Therefore, we have reasonable grounds to believe that the normal activities should be compact in the feature space. Consequently, we use batch clustering to cluster normal video segments to enhance the intra-class compactness of normal features. A larger inter-class distance in abnormal video indicates that normal and abnormal are separated by a higher probability. Therefore, we perform batch clustering on abnormal video segments to enhance the inter-class dispersion of normal and abnormal features.

Here, we propose batch clustering to provide supervision to enhance the discriminative power of features as shown in Figure \ref{fig:model}. For all normal videos in a batch, the feature representations of each video are clustered into two clusters. Since all segments in normal videos are normal, we try to close the distance of the two clusters to ensure the intra-class compactness of the normal features. All the abnormal videos in a batch are also clustered into two clusters. Since there are both abnormal and normal video segments in the abnormal videos, we try to push the centers of the two clusters away from each other to achieve the inter-class dispersion of normal and abnormal features. Specifically, for the abnormal or normal videos in a batch, we use the K-means algorithm to cluster the normalized intermediate feature representations, which is the output of the first layer of GCN. The loss based on batch clustering is shown below:
\begin{equation}
	L_{b c}=\left\{\begin{array}{c}
		\min (d, \mu), \text { if }\left\{V_{i}\right\}_{i=1}^{b} \text { are normal videos } \\
		\frac{1}{d}, \text { if }\left\{V_{i}\right\}_{i=1}^{b} \text { are abnormal videos }
	\end{array}\right.
\end{equation}
where $d=c_{1}-c_{2}$ is the distance between two cluster centers, and $c_{1}$, $c_{2}$ are the two cluster centers. $\mu$ is an upper bound that helps the model to be robust to different videos, and $b$ is the batch size.

We train the model using the $k$-max loss function to expand the inter-class distance between abnormal and normal segments, denoted as follows:
\begin{equation}
	L_{k-\max }=-\frac{1}{k_{i}} \sum_{s_{i, j} \in p_{i}}\left[y_{i} \log \left(s_{i, j}\right)+(1-y_{i}) \log \left(1-s_{i, j}\right)\right]
\end{equation}
where $p_i$ is the first $k_i$ large elements in $S_i$ of video $V_i$, $k_{i}=\left\lfloor\frac{T_{i}}{8}+1\right\rfloor$, $y_i$ is the video-level label of video $V_i$. So, the total loss function is expressed as:

\begin{equation}\label{loss2}
	L=L_{k-\max }+\lambda_{1} L_{b c}
\end{equation}
where $\lambda_{1}$ is a trade-off hyperparameter to keep the balance between two losses.

\subsection{Cross-batch learning strategy}
\label{batch}
The data in WSVAD task is highly imbalanced. However, only a  mini-batch of samples can be accessed in each iteration. The performance of anomaly detection models can be improved when the batch size becomes larger on large-scale datasets. However, simply scaling up the batch is not an ideal solution because batch size is limited by GPU memory and computational cost. A simple solution to collect rich information is to introduce information from previous batches at each training iteration to enable the model to better understand the underlying distribution of the data. Thus, video segments from past batches can also serve as an important reference when performing $K$-means clustering on video segments of the current batch. Previously, we perform batch clustering based on $K$-means by directly selecting any two video segments from all video segments as the initial clustering centers, but the final clustering results of the $K$-means depend on the selection of the initial clustering centers heavily. In order to select the most appropriate initial clustering centers, we introduce the clustering results of previous batches to provide guidance for the current batch clustering, which can help model training.

Specifically, at the $t$-th epoch, $C^a$ and $C^n$ are constructed in the iteration process to store the learned knowledge. That is, during the $i$-th iteration, we cluster all abnormal video segments in a batch, and add the two clustering centers into $C^a$. We will obtain $C^{a}=\left\{c_{i, 1}^{a}, c_{i, 2}^{a}\right\}_{i=1}^{m}$ at the end of the $t$-th epoch, where $m$ is the number of iterations in an epoch. We also do the above operation for all normal video segments in a batch to get $C^{n}=\left\{c_{i, 1}^{n}, c_{i, 2}^{n}\right\}_{i=1}^{m}$.

At the $(t+1)$-th epoch, we use the stored information to guide the batch clustering. We cluster the data in $C^a$ to obtain the clustering centers $c_1^a$, $c_2^a$, which are used as the initial clustering centers for the batch clustering of all abnormal video segments in each batch of the current epoch. Meanwhile, the data in $C^n$ are clustered to obtain the clustering centers $c_1^n$, $c_2^n$, which are used as the initial clustering centers for all normal video segments in each batch of the current epoch. Dynamically updating the data in $C^a$ and $C^n$ among different epochs and introducing them into the batch clustering as a priori information can improve the effect of batch clustering and further improve the performance of the anomaly detection task.

\subsection{Anomaly score generation based on batch clustering guidance}
When the available labels are limited, the labeled samples often do not provide sufficient supervised information for the model, so the deep model is prone to overfitting. Since WSVAD only considers video-level labels, the knowledge that the model can learn will be limited. To address this problem, we use cluster labels obtained from batch clustering to guide backbone network to predict the anomaly scores of segments. The unsupervised information is effectively transferred to the weakly supervised learning process to improve the performance of the anomaly detection task.

For the labeled normal videos, each segment of these videos can simply be annotated as normal because there are no abnormal events in it. However, in the case of abnormal videos, there are also some normal events, so we use batch clustering results to generate pseudo label for each segment of the abnormal video. All segments of abnormal videos are clustered into two clusters assuming that one cluster would contain normal segments, while the other would contain abnormal. Therefore, we need to analyze which of the two clusters contains mostly normal segments and which contains mostly abnormal segments so that we can assign the appropriate pseudo labels for video segments. We obtain the pseudo labels of the segments by the similarity score between the anomaly scores and the clustering labels. Specifically, we compute the cosine similarity score $S_1$ between the anomaly score $s_i\in[0,1]$ of video $V_i$ predicted by the backbone network and the label $y_{i}^{c} \in\{0,1\}$ generated by clustering, and the cosine similarity score $S_2$ between $s_i$ and the inverted clustering label $\neg y_{i}^{c}$. Finally, the pseudo label $y_{i, j}^{p}$ of the $j$-th segment of video $V_i$ is given by the following equation:
\begin{equation}
	y_{i, j}^{p}=\left\{\begin{array}{cc}
		y_{i, j}^{c}, & \text { if } s_{1}>s_{2} \\
		\neg y_{i, j}^{c}, & \text { otherwise }
	\end{array}\right.
\end{equation}

The pseudo labels generated by batch clustering for video segments and the anomaly scores generated by adaptive graph convolutional networks for video segments can complement each other in the anomaly detection task. In an anomalous video, if the pseudo label of a video segment is 1, the segment has a high probability of being abnormal. Therefore, we expand the anomaly score of the segments with a pseudo label 1 in the abnormal video. In particular, in an abnormal video, if the pseudo label of the video segment is 1, the abnormal score of the video segment becomes $\min \left(\alpha \times s_{i, j}, 1\right)$; if the pseudo label of the video segment is 0, the abnormal score of the video segment remains unchanged. The anomaly score of each video segment is given by the following equation:
\begin{equation}
	s_{i, j}=\left\{\begin{array}{c}
		\min \left(\alpha \times \tilde{s}_{i, j}, 1\right), \text { if } y_{i}=1 \text { and } y_{i, j}^{p}=1 \\
		\tilde{s}_{i, j}, \text { otherwise }
	\end{array}\right.
\end{equation}
where $\alpha$ is the expansion factor.
		
\section{Experimental results}

\subsection{Datasets and evaluation metrics}
UCF-Crime \cite{sultani2018real} is a large-scale dataset which spans over 128 hours of surveillance videos. It covers 13 realistic anomalies. The entire dataset contains 1,900 long untrimmed videos, of which 1610 videos with video-level label are used for training and the rest for testing.

ShanghaiTech \cite{luo2017revisit} is a medium-scale campus surveillance dataset containing 437 videos. Following Zhong et al. \cite{zhong2019graph}, we split the data into two subsets: the training set consisting of 175 normal videos and 63 anomalous videos, and the test set containing 155 normal videos and 44 anomalous videos.

Following previous work \cite{sultani2018real}, we use the area under curve (AUC) of the receiver operating characteristic curve (ROC) at the frame level as the criterion for the model, and a higher AUC value indicates a better detection of the model.
\begin{table}[tp]
	\caption{Frame-level AUC performance comparisons.}
	\label{table_1}
	\begin{tabular}{cccc}
		\hline
		Model & Feature & UCF-Crime & ShanghaiTech \\
		\hline
		Sultani et al. \cite{sultani2018real} & I3D RGB  & 76.92 & 86.30 \\
		Zhong et al. \cite{zhong2019graph} & C3D RGB & 81.08 & 76.44\\
		AR\_Net \cite{wan2020weakly} & I3D RGB  & 78.96 & 85.38\\
		SRF \cite{zaheer2020self} & I3D RGB  & 79.54 & 84.17\\
		Wu et al. \cite{wu2020not} & I3D RGB  & 82.44 & / \\
		CLAWS \cite{zaheer2020claws} & C3D RGB  & 83.03 & 89.67\\
		MIST \cite{feng2021mist} & I3D RGB  & 82.30 & 94.83\\
		BN-SVP \cite{sapkota2022bayesian} & I3D RGB  & 83.39 & 96.00\\
		MCR \cite{gong2022multi} &I3D RGB & 81.0 & 90.10\\
		MSLNet \cite{li2022self} & I3D RGB  & 85.30 & 96.08\\
		Ours & I3D RGB  & \pmb{85.87} & \pmb{96.45}\\
		\hline
	\end{tabular}
\end{table}
\begin{table}[t]
	\caption{Ablation study on UCF-Crime dataset($L_{bc}$: batch clustering based loss, CBL: cross-batch learning strategy, BCG: anomaly score generation based on batch clustering guidance).}
	\label{bottomup1}
	\centering
	\begin{tabular}{ccccc}
		\hline
		backbone & $L_{bc}$ & CBL & BGG & AUC(\%) \\
		\hline
		$\checkmark$ & & & & 84.67\\
		$\checkmark$ & $\checkmark$ & & & 85.21 \\
		$\checkmark$ & $\checkmark$ & $\checkmark$ & & 85.56 \\
		$\checkmark$ & $\checkmark$ & $\checkmark$ & $\checkmark$ & 85.87 \\
		\hline
	\end{tabular}
\end{table}
\begin{table}[t]
	\begin{center}
		\caption{Performance comparison of batch clustering with different network layer outputs on UCF-Crime.}
		\label{layer}
		\begin{tabular}{cc}
			\hline
			clustering layer & AUC(\%) \\
			\hline
			FC & 84.96\\
			GCN-1 & 85.21 \\
			GCN-2 & 84.27\\
			\hline
		\end{tabular}
	\end{center}
\end{table}
\begin{table}[t]
	\begin{center}
		\caption{Performance comparison of different cross-batch learning strategies on UCF-Crime.}
		\label{cross_batch}		
		\begin{tabular}{cc}
			\hline
			different strategies & AUC(\%) \\
			\hline
			way1 & 85.56\\
			way2 & 85.42\\
			way3 & 85.51\\
			way4 & 85.34 \\
			\hline
		\end{tabular}
	\end{center}
\end{table}
\begin{figure}[t]
	\centering
	\includegraphics[width=0.9\linewidth]{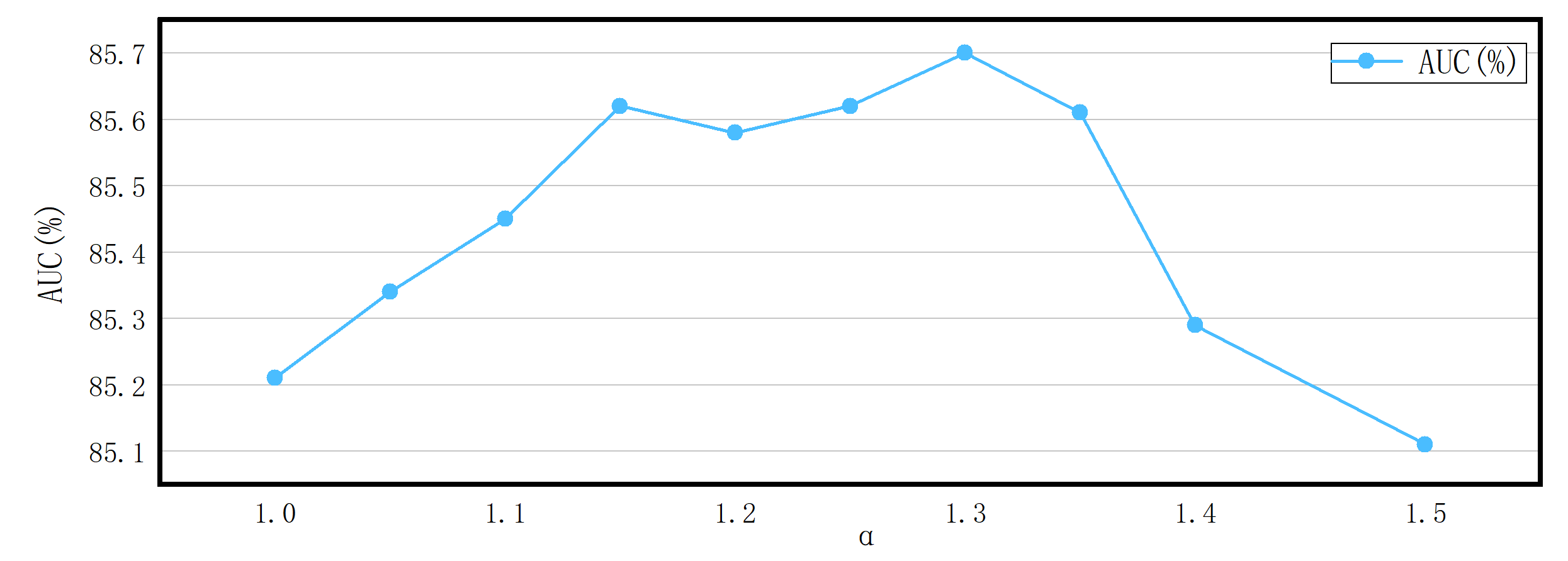}
	\caption{Performance comparison with different expansion factors $\alpha$ on UCF-Crime.}
	\label{fig:expand}
\end{figure}
\begin{figure*}[t]
	\begin{minipage}[b]{0.245\linewidth}
		\centerline{\epsfig{figure=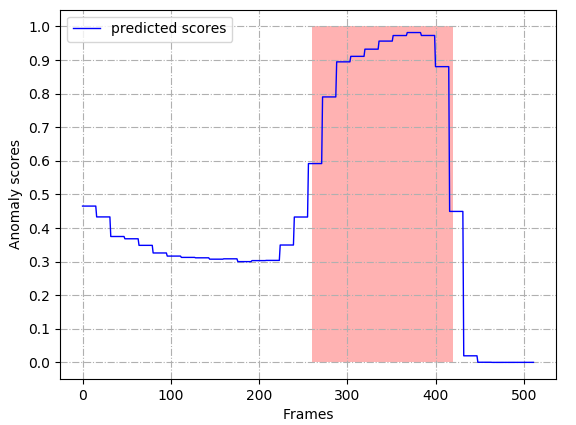,width=4.0cm,height=2.5cm}}
		\vspace{-0.1cm}
		\centerline{(a) Explosion025}\medskip
		\label{fig:res1}
	\end{minipage}
	\hfill
	\begin{minipage}[b]{0.245\linewidth}\label{fig:res2}
		\centerline{\epsfig{figure=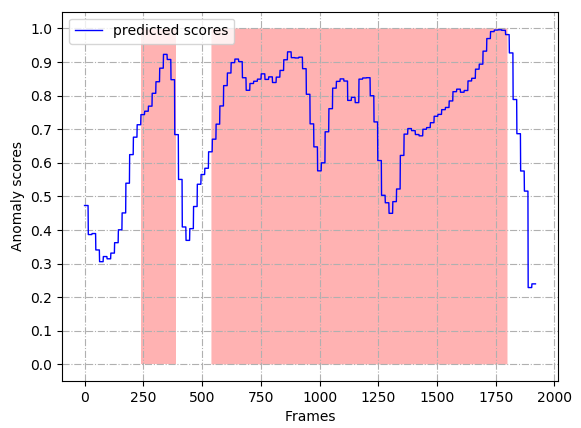,width=4.0cm,height=2.5cm}}
		\vspace{-0.1cm}
		\centerline{(b) Burglary037}\medskip
	\end{minipage}
	\hfill
	\begin{minipage}[b]{0.245\linewidth}\label{fig:res3}
		\centerline{\epsfig{figure=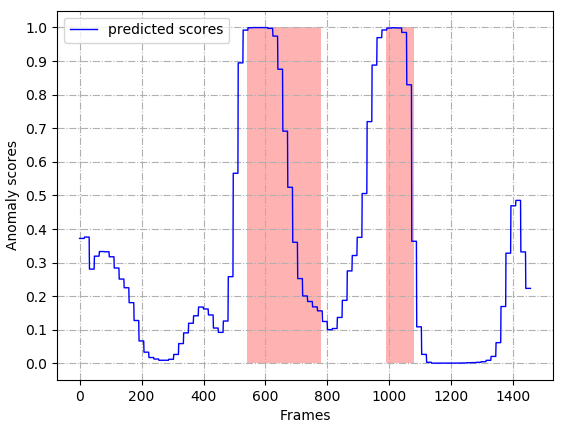,width=4.0cm,height=2.5cm}}
		\vspace{-0.1cm}
		\centerline{(c) Vandalism036}\medskip
	\end{minipage}
	\hfill
	\begin{minipage}[b]{0.245\linewidth}
		\centerline{\epsfig{figure=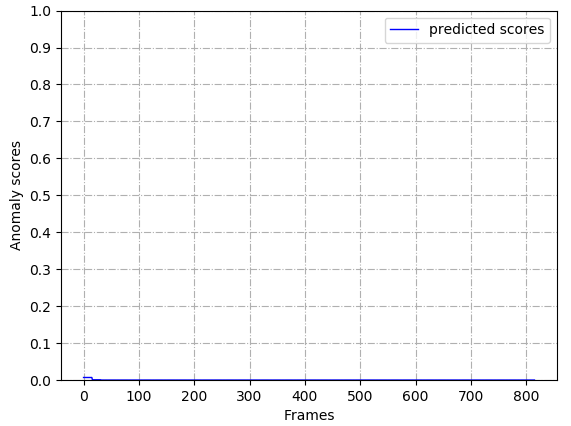,width=4.0cm,height=2.5cm}}
		\vspace{-0.1cm}
		\centerline{(d) Normal939}\medskip
	\end{minipage}
	
	\caption{Visualization of the testing results on UCF-Crime. Red regions are ground truths of anomalous events.}
	\label{fig:res}
\end{figure*}
\subsection{Experimental details}
We extract the 2048-dimensional features from the “mix 5c” layer of I3D \cite{carreira2017quo}. Following previous work \cite{wu2020not}, we extract $T$ segments from the video uniformly to represent the whole video. By default, we set $T$ to 150 for UCF-Crime and 100 for ShanghaiTech. The fully connected layer in the model has 512 nodes, and the graph convolutional network layer has 128, 32 and 1 nodes, respectively. Our model is trained with a mini-batch size of 64 using the Adam optimizer. For hyper-parameters, $\alpha$ is set to 1.3.

\subsection{Experimental results}
We compare our method with the current methods on two datasets, and show results in Table \ref{table_1}. For the UCF-Crime dataset, comparison results show that our method outperforms all comparison methods. Using the same I3D-RGB features, our method outperforms the previous GCN-based method by 4.79\% over Zhong et al. \cite{zhong2019graph} and 3.43\% over Wu et al. \cite{wu2020not}. Our method also surpasses previous methods using clustering by 6.33\% over SRF \cite{zaheer2020self} and 2.84\% over CLAWS \cite{zaheer2020claws}. It also achieves better performance compared to previous weakly supervised methods on the ShanghaiTech dataset. It is 20.01\% higher than the GCN-based method \cite{zhong2019graph}, 12.29\% and 6.78\% higher than the clustering-based method \cite{zaheer2020self} and \cite{zaheer2020claws}, respectively.

\subsection{Ablation experiments}
We perform ablation study to investigate the contribution of each component of our proposed method. We start by evaluating only the backbone network and observe the performance gain while adding different modules. The result on UCF-Crime is shown in Table \ref{bottomup1}. The backbone network achieves 84.67\% AUC while the addition of batch clustering based loss improves the performance to 85.21\% and the addition of cross-batch learning strategy further improves the performance to 85.56\% which validates their effectiveness. Addition of the anomaly score generation based on batch clustering guidance improves the performance to 85.87\%, demonstrating that the pseudo labels generated by batch clustering can guide the backbone network.

We select different network layer outputs for batch clustering to investigate their effects on model performance. We choose the outputs of FC layer (FC), the first GCN layer (GCN-1), and the second GCN layer (GCN-2), respectively. As shown in Table \ref{layer}, using GCN-1 achieves the best result with 0.25\% higher than that using FC, probably because GCN-1 exploits the temporal relationship between video segments. The performance using GCN-2 is 0.94\% lower than that using GCN-1 because GCN-1 contains more information due to its higher dimensionality compared to GCN-2.

We conduct experiments and show comparison results to evaluate the effects of different cross-batch learning strategies. The first method is the one mentioned in Section \ref{batch}. The second method is to save the clustering centers obtained from the previous batch clustering as the initial clustering centers when clustering the current batch. The third method is to save the clustering centers obtained from all previous batch clustering and use the clustering centers obtained by clustering them again as the initial clustering centers for the current batch clustering. The fourth method is to save the clustering centers of all the batches from the previous epoch as clustering samples to participate in the clustering of each batch in the current epoch. The Table \ref{cross_batch} shows that all our proposed cross-batch learning strategies improve the model performance. The first method has the best performance, and the fourth method performing less well, with only a small improvement.

We also conduct experiment to further investigate the expansion factor $\alpha$, and show the change of performance with different expansion factors $\alpha$ on UCF-Crime in Figure \ref{fig:expand}. We can observe that the AUC first increases and then decreases as the value of $\alpha$ increases. The model achieves the best performance when $\alpha$ is 1.3.
\subsection{Qualitative Result and Analysis}
To further evaluate the performance of our method, we visualize the anomaly score curves, as shown in Figure \ref{fig:res}. The figure shows the ground truth and the predicted anomaly scores of three abnormal videos and one normal video from UCF-Crime. It is obviously that our model exactly localizes the anomalous events, showing the effectiveness and robustness of our model. Our method successfully predicts both short-term anomalous events and long-term anomalous events. In addition, our method can also detect multiple anomalous events in a video.

\section{Conclusion}
In this work, we propose a novel WSVAD model based on cross-batch clustering guidance. The method enhances feature discrimination by binary batch clustering for normal and abnormal videos within a batch separately. In addition, a cross-batch learning strategy is incorporated to solve the data imbalance problem caused by the mini-batch training strategy, allowing the model to better capture the potential distribution of the data. Finally, the pseudo labels generated by batch clustering guide the backbone network to generate the anomaly scores, which further enhances the separability of normal and anomalous. The experimental results show that the proposed method achieves significant improvements on commonly-used WSVAD datasets.

\bibliographystyle{IEEEtran}
\bibliography{ref}

\begin{thebibliography}{10}
\providecommand{\url}[1]{#1}
\csname url@samestyle\endcsname
\providecommand{\newblock}{\relax}
\providecommand{\bibinfo}[2]{#2}
\providecommand{\BIBentrySTDinterwordspacing}{\spaceskip=0pt\relax}
\providecommand{\BIBentryALTinterwordstretchfactor}{4}
\providecommand{\BIBentryALTinterwordspacing}{\spaceskip=\fontdimen2\font plus
\BIBentryALTinterwordstretchfactor\fontdimen3\font minus
  \fontdimen4\font\relax}
\providecommand{\BIBforeignlanguage}[2]{{%
\expandafter\ifx\csname l@#1\endcsname\relax
\typeout{** WARNING: IEEEtran.bst: No hyphenation pattern has been}%
\typeout{** loaded for the language `#1'. Using the pattern for}%
\typeout{** the default language instead.}%
\else
\language=\csname l@#1\endcsname
\fi
#2}}
\providecommand{\BIBdecl}{\relax}
\BIBdecl

\bibitem{sultani2018real}
W.~Sultani, C.~Chen, and M.~Shah, ``Real-world anomaly detection in
  surveillance videos,'' in \emph{Proc. IEEE Conf. Comput. Vision Pattern
  Recognit.}, 2018, pp. 6479--6488.

\bibitem{wan2020weakly}
Y.~Wan, B., X.~Xia, and J.~Mei, ``Weakly supervised video anomaly detection via
  center-guided discriminative learning,'' in \emph{Proc. IEEE Int. Conf.
  Multimedia Expo.}, 2020, pp. 1--6.

\bibitem{li2022self}
S.~Li, F.~Liu, and L.~Jiao, ``Self-training multi-sequence learning with
  transformer for weakly supervised video anomaly detection,''
  \emph{Proceedings of the AAAI, Virtual}, vol.~24, 2022.

\bibitem{gong2022multi}
Y.~Gong, C.~Wang, X.~Dai, S.~Yu, L.~Xiang, and J.~Wu, ``Multi-scale
  continuity-aware refinement network for weakly supervised video anomaly
  detection,'' in \emph{2022 IEEE International Conference on Multimedia and
  Expo}, 2022, pp. 1--6.

\bibitem{zaheer2020self}
M.~Z. Zaheer, A.~Mahmood, H.~Shin, and S.-I. Lee, ``A self-reasoning framework
  for anomaly detection using video-level labels,'' \emph{IEEE Signal Process.
  Lett.}, vol.~27, pp. 1705--1709, 2020.

\bibitem{zaheer2020claws}
M.~Z. Zaheer, A.~Mahmood, M.~Astrid, and S.-I. Lee, ``Claws: Clustering
  assisted weakly supervised learning with normalcy suppression for anomalous
  event detection,'' in \emph{Proc. Eur. Conf. Comput. Vis.}, 2020, pp.
  358--376.

\bibitem{wang2020cross}
X.~Wang, H.~Zhang, W.~Huang, and M.~R. Scott, ``Cross-batch memory for
  embedding learning,'' in \emph{Proc. IEEE Conf. Comput. Vision Pattern
  Recognit.}, 2020, pp. 6388--6397.

\bibitem{wu2019long}
C.-Y. Wu, C.~Feichtenhofer, H.~Fan, K.~He, P.~Krahenbuhl, and R.~Girshick,
  ``Long-term feature banks for detailed video understanding,'' in \emph{Proc.
  IEEE Conf. Comput. Vision Pattern Recognit.}, 2019, pp. 284--293.

\bibitem{cao2022adaptive}
C.~Cao, X.~Zhang, S.~Zhang, P.~Wang, and Y.~Zhang, ``Adaptive graph
  convolutional networks for weakly supervised anomaly detection in videos,''
  \emph{arXiv preprint arXiv:2202.06503}, 2022.

\bibitem{carreira2017quo}
J.~Carreira and A.~Zisserman, ``Quo vadis, action recognition? a new model and
  the kinetics dataset,'' in \emph{Proc. IEEE Conf. Comput. Vision Pattern
  Recognit.}, 2017, pp. 6299--6308.

\bibitem{perera2019learning}
P.~Perera and V.~M. Patel, ``Learning deep features for one-class
  classification,'' \emph{IEEE Transactions on Image Processing}, vol.~28,
  no.~11, pp. 5450--5463, 2019.

\bibitem{luo2017revisit}
W.~Luo, W.~Liu, and S.~Gao, ``A revisit of sparse coding based anomaly
  detection in stacked rnn framework,'' in \emph{Proceedings of the IEEE
  international conference on computer vision}, 2017, pp. 341--349.

\bibitem{zhong2019graph}
J.-X. Zhong, N.~Li, W.~Kong, S.~Liu, T.~H. Li, and G.~Li, ``Graph convolutional
  label noise cleaner: Train a plug-and-play action classifier for anomaly
  detection,'' in \emph{Proc. IEEE Conf. Comput. Vision Pattern Recognit.},
  2019, pp. 1237--1246.

\bibitem{wu2020not}
P.~Wu, J.~Liu, Y.~Shi, Y.~Sun, F.~Shao, Z.~Wu, and Z.~Yang, ``Not only look,
  but also listen: Learning multimodal violence detection under weak
  supervision,'' in \emph{Proc. Eur. Conf. Comput. Vis.}\hskip 1em plus 0.5em
  minus 0.4em\relax Springer, 2020, pp. 322--339.

\bibitem{feng2021mist}
J.-C. Feng, F.-T. Hong, and W.-S. Zheng, ``Mist: Multiple instance
  self-training framework for video anomaly detection,'' in \emph{Proc. IEEE
  Conf. Comput. Vision Pattern Recognit.}, 2021, pp. 14\,009--14\,018.

\bibitem{sapkota2022bayesian}
H.~Sapkota and Q.~Yu, ``Bayesian nonparametric submodular video partition for
  robust anomaly detection,'' in \emph{Proc. IEEE Conf. Comput. Vision Pattern
  Recognit.}, 2022, pp. 3212--3221.

\end{thebibliography}

\end{document}